%% file: root.tex
\documentclass[letterpaper, 10 pt, journal, twoside]{IEEEtran}
\usepackage{amsmath,amsfonts}

\usepackage[utf8]{inputenc}
\usepackage{multirow}
\usepackage{csquotes}
\usepackage{slashbox}
\usepackage{algorithm2e}
\usepackage{enumitem}
\usepackage{array}
\usepackage{adjustbox}
\usepackage{booktabs}
\usepackage{subcaption}
\usepackage[font=small,tableposition=top]{caption}
\usepackage{xcolor}
\usepackage{url}
\usepackage{graphicx}
\usepackage{cite}
\usepackage{collectbox}
\usepackage{multirow}
\RestyleAlgo{ruled}
\usepackage{etoolbox}
\usepackage{cancel}
\makeatletter
\let\NAT@parse\undefined
\makeatother

\usepackage[colorlinks=false]{hyperref}

\definecolor{mygreen}{rgb}{0.0, 0.5, 0.0}
\definecolor{mymagenta}{rgb}{0.75, 0.0, 0.75}
\definecolor{myorange}{rgb}{1.0, 0.6470588235294118, 0.0}
\begin{document}

\title{MoVEInt: Mixture of Variational Experts for Learning Human-Robot Interactions from Demonstrations}

\author{Vignesh Prasad$^1$, Alap Kshirsagar$^1$, Dorothea Koert$^{1,3}$\\Ruth Stock-Homburg$^2$, Jan Peters$^{1,3,4,5}$, Georgia Chalvatzaki$^{1,5}$\vspace{-2em}% <-this % stops a space
\thanks{%Manuscript received: November 30, 2023; Revised March 5, 2024; Accepted April 3, 2024. This paper was recommended for publication by Editor Aleksandra Faust upon evaluation of the Associate Editor and Reviewers' comments. 
This research work has received funding from the German Research Foundation (DFG) Emmy Noether Programme (CH 2676/1-1), the German Federal Ministry of Education and Research (BMBF) Projects \enquote{IKIDA} (Grant no.: 01IS20045) and \enquote{KompAKI} (Grant no.: 02L19C150), the Excellence Program, \enquote{The Adaptive Mind}, of the Hessian Ministry of Higher Education, Science, Research and Art, the EU’s Horizon Europe projects \enquote{MANiBOT} (Grant no.: 101120823) and \enquote{ARISE} (Grant no.: 101135959).}%Use only for final RAL version
\thanks{\hspace{-1em}$^1$ Department of Computer Science, TU Darmstadt, Germany}%
\thanks{\hspace{-1em}$^2$ Chair for Marketing and Human Resource Management (MuP), Department of Law and Economics, TU Darmstadt, Germany}%
\thanks{\hspace{-1em}$^3$ Centre for Cognitive Science, TU Darmstadt, Germany}%
\thanks{\hspace{-1em}$^4$ German Research Center for AI (DFKI), Darmstadt, Germany}%
\thanks{\hspace{-1em}$^5$ Hessian Center for Artificial Intelligence (hessian.AI), Darmstadt, Germany}%
% $^7$ Interactive Robot Perception and Learning group (PEARL), Department of Computer Science, TU Darmstadt, Germany.\newline
% $^1$ Institute for Intelligent Autonomous Systems (IAS), Department of Computer Science, TU Darmstadt, Germany.\newline
% $^2$ Chair for Marketing and Human Resource Management (MuP), Department of Law and Economics, TU Darmstadt, Germany.\newline
% $^3$ Interactive AI Algorithms \& Cognitive Models for Human-AI Interaction (IKIDA), Department of Computer Science, TU Darmstadt, Germany.\newline
% $^4$ Centre for Cognitive Science, TU Darmstadt, Germany.\newline
% $^5$ Systems AI for Robot Learning (SAIROL), German Research Center for AI (DFKI), Darmstadt, Germany.\newline
% $^6$ Hessian Center for Artificial Intelligence, Darmstadt, Germany\newline
% $^7$ Interactive Robot Perception and Learning group (PEARL), Department of Computer Science, TU Darmstadt, Germany.\newline
% $^8$ Center for Mind, Brain \& Behavior (CMBB), Uni. Marburg \& JLU Giessen, Germany
% }% <-this % stops a space
\thanks{Digital Object Identifier (DOI): \href{https://doi.org/10.1109/LRA.2024.3396074}{10.1109/LRA.2024.3396074}}
\thanks{Email: ({\tt\footnotesize \href{mailto:vignesh.prasad@tu-darmstadt.de}{vignesh.prasad@tu-darmstadt.de}})}
\thanks{Additional resources: \url{https://bit.ly/MoVEInt}.}
}

\markboth{IEEE Robotics and Automation Letters. Preprint Version. Accepted April, 2024}
{Prasad \MakeLowercase{\textit{et al.}}: M\MakeLowercase{o}VEI\MakeLowercase{nt}: Mixture of Variational Experts for Learning Human-Robot Interactions from Demonstrations} 

\maketitle
% \thispagestyle{empty}
% \pagestyle{empty}

%%%%%%%%%%%%%%%%%%%%%%%%%%%%%%%%%%%%%%%%%%%%%%%%%%%%%%%%%%%%%%%%%%%%%%%%%%%%%%%%
\begin{abstract}

% In this paper, we explore learning modular latent space policies for Human-Robot Interaction. 
% In this work, we explore learning latent policy mixtures to model interaction dynamics for Human-Robot Interaction (HRI) from demonstrations. We show how to learn shared interaction dynamics in a modular manner by leveraging Mixture Density Networks (MDNs) to model human conditioned priors that learn a mixture of latent policies for reactively generating robot actions. We show how our formulation manifests itself as inferring a joint distribution over the actions of the human and the robot using Gaussian Mixture Regression. We show through extensive experiments on a variety of datasets that our approach generates highly accurate robot motions as compared to previous HMM-based or recurrent approaches for learning HRI. We further validate this with real world robot experiments showing the efficacy of our approach.
Shared dynamics models are important for capturing the complexity and variability inherent in Human-Robot Interaction (HRI). Therefore, learning such shared dynamics models can enhance coordination and adaptability to enable successful reactive interactions with a human partner. In this work, we propose a novel approach for learning a shared latent space representation for HRIs from demonstrations in a Mixture of Experts fashion for reactively generating robot actions from human observations. We train a Variational Autoencoder (VAE) to learn robot motions regularized using an informative latent space prior that captures the multimodality of the human observations via a Mixture Density Network (MDN). 
We show how our formulation derives from a Gaussian Mixture Regression formulation that is typically used approaches for learning HRI from demonstrations such as using an HMM/GMM for learning a joint distribution over the actions of the human and the robot. We further incorporate an additional regularization to prevent \enquote{mode collapse}, a common phenomenon when using latent space mixture models with VAEs. We find that our approach of using an informative MDN prior from human observations for a VAE generates more accurate robot motions compared to previous HMM-based or recurrent approaches of learning shared latent representations, which we validate on various HRI datasets involving interactions such as handshakes, fistbumps, waving, and handovers. Further experiments in a real-world human-to-robot handover scenario show the efficacy of our approach for generating successful interactions with four different human interaction partners.

\end{abstract}
% \begin{keywords}

%%%%%%%%%%%%%%%%%%%%%%%%%%%%%%%%%%%%%%%%%%%%%%%%%%%%%%%%%%%%%%%%%%%%%%%%%%%%%%%%
\section{Introduction}

Ensuring a timely response while performing an interaction can enable a feeling of connectedness to ones partner~\cite{marsh2009social}. Therefore, learning to generate response motions for a robot in a timely manner for Human-Robot Interaction (HRI) is an important aspect of the interaction. One way to do so is by learning a shared representation space between the human and the robot~\cite{butepage2020imitating,jonnavittula2021learning,joshi2020electric,prasad2022mild,prasad2023learning,vogt2017system,wang2022co}. An important aspect of such approaches, for learning HRI from demonstrations, is accurately capturing the multimodality of the underlying data to effectively capture the underlying skills and accurately generate response motions.

\begin{figure}[t!]
    \centering
    % \includegraphics[width=\linewidth]{teaser}
    % \sqbox{Teaser Image}
    % \begin{subfigure}[b]{0.44\linewidth}
    %     \includegraphics[width=\linewidth]{img/kobo-recording/kobo-rviz-frame3.png}
    %     \caption{}
    % \end{subfigure} \begin{subfigure}[b]{0.44\linewidth}
    %     \includegraphics[width=\linewidth]{img/pepper-recording/pepper-rviz-frame3.png}
    %     \caption{}
    % \end{subfigure}
    \includegraphics[width=0.8\linewidth]{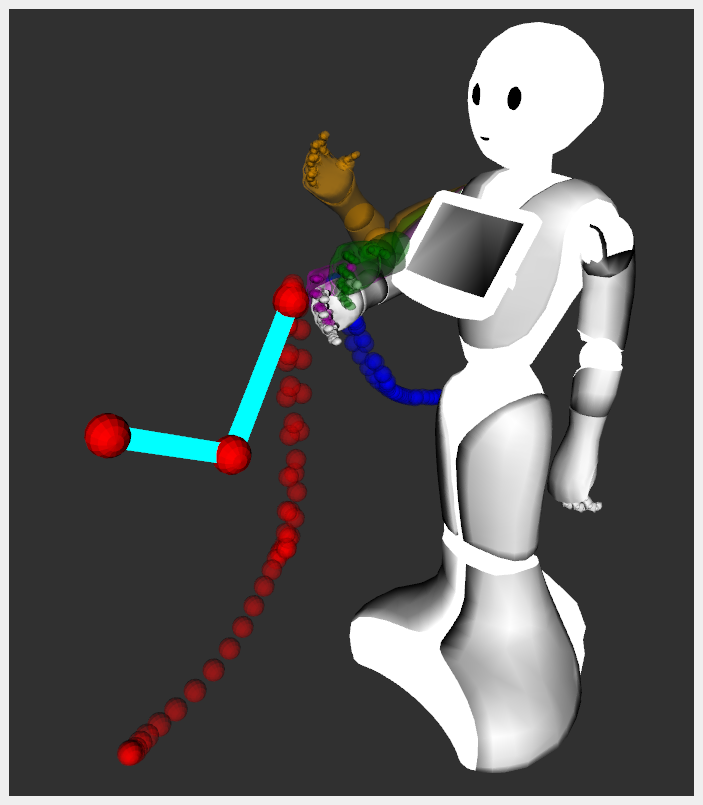}
    \caption{Target poses generated in a reactive manner by the mixture of policies learned by MoVEInt for a Handshake interaction with the humanoid robot Pepper. MoVEInt generates multiple policies (shown in \textcolor{mygreen}{green}, \textcolor{mymagenta}{magenta}, and \textcolor{myorange}{orange}) based on human observations which are combined to generate suitable robot motions.}
    \label{fig:teaser}
    \vspace{-2em}
\end{figure}

Vogt et al.~\cite{vogt2017system} showed the use of shared latent spaces in an Imitation Learning (IL) approach by decomposing interactions into multiple segments in a lower dimensional space with an underlying Gaussian Mixture Model (GMM) and learning the sequence of key poses using a Hidden Markov Model (HMM) to define the progression of an interaction. %The HMM-GMM model is then used to define constraints to optimize a mesh topology that generates the robot motions in an online manner. 
In our previous work, MILD~\cite{prasad2022mild,prasad2023learning}, we explore learning a shared latent space model using a Variational Autoencoder (VAE) wherein we learn a joint distribution over the trajectories of both the human and the robot using an HMM with underlying Gaussian States to represent the multimodality of the demonstrations. Rather than extracting key poses as in~\cite{vogt2017system}, we generate the robot's motion using Gaussian Mixture Regression (GMR) from the underlying HMM based on the human's observations in a reactive manner. In doing so, we achieve better accuracy than using a recurrent representation of the shared latent dynamics~\cite{butepage2020imitating}. 

Generative approaches have been used to jointly learn human and robot policies for collaborative tasks~\cite{wang2022co,tormos2023explainable,sengadu2023dec} and discover different underlying latent \enquote{strategies} of the human but only for a given task. In our work, we further explore how underlying latent strategies can be learned from different tasks in a dataset rather than just a single task by using a mixture distribution to predict different latent policies, which are then combined in a Mixture of Experts fashion. An example of this can be seen in Fig.~\ref{fig:teaser}, where we show a handshake interaction with the Pepper robot. Trained on a dataset of different physical interactions like waving, handshakes and fistbumps, we see the different policies that get predicted (the reconstructions of which are shown by the different colored arms of Pepper) which are then combined in the latent space yielding a suitable response motion. 

% However, in this work, we enforce a more explicit structure to learn different strategies

% In this paper, we explore a similar approach with a key distinction that we enforce an explicit mixture distribution to predict different latent policies which are then blended together in a Mixture of Experts fashion.  we consider the human as part of the environment state that the robot reacts to, instead of predicting the human's actions as a separate policy as done in~\cite{wang2022co}. Moreover, instead of an implicit extraction of different \enquote{strategies} of the human, we explicitly show how such an imitation learning approach can further be decomposed to enable a modular approach that can conditionally generate the robot motions~\cite{vogt2017system,prasad2022mild}. 

% \cite{butepage2020imitating} similarly demonstrate learning shared latent representations for learning HRI dynamics from Human Demonstrations using Behavior Cloning. In our previous work \cite{prasad2022mild}, we extend the idea of learning shared representations presented in \cite{butepage2020imitating} by using Hidden Markov Models with underlying Gaussian states to better capture the multimodal representations of the shared dynamics leading to more accurate interactions.
% Transition sentence for going from shared latent representations (or explicitly from Co-GAIL) to using MDNs for learning modular/multimodal policies.

To learn multiple latent policies and effectively combine them, we employ Mixture Density Networks (MDNs)~\cite{bishop1994mixture} to capture the multimodality of the demonstrations. MDNs predict a mixture of Gaussians and the corresponding mixture coefficients yielding a multimodal prediction, rather than a unimodal distribution or a single output. MDN policy representations have been widely used in Imitation Learning and Reinforcement Learning (RL) on a variety of tasks such as autonomous driving~\cite{baheri2022safe,kuutti2021adversarial}, entropy regularization for imitation learning~\cite{lee2018maximum}, adapting to multiple goals~\cite{zhou2020movement}, incorporating human intentions for robot tasks~\cite{thabet2019sample}, predicting gaze behaviors for robot manipulation~\cite{kim2022memory} or more generally as improved attention mechanisms~\cite{bazzani2016recurrent,abolghasemi2019pay} or as forward models of the environment~\cite{ha2018recurrent,schrempf2005generic,zhang2018auto,rahmatizadeh2018virtual}. %Furthermore, Recurrent MDNs have shown to

With these key ideas for learning shared multimodal latent policy representations for HRI, the main contributions of our paper are as follows. 
% \begin{enumerate}
    We propose \enquote{MoVEInt}, a novel framework that employs a Mixture of Variational Experts for learning Human-Robot Interactions from demonstrations through a shared latent representation of a human and a robot. We learn latent space policies in a Mixture of Experts fashion via a Mixture Density Network (MDN) to encode the latent trajectory of a human partner, regularize the robot embeddings, and subsequently, predict the robot motions reactively. 
    We show how our proposed formulation extends from Gaussian Mixture Regression (GMR) which is typically used in HMM-based approaches for learning HRI from demonstrations. We leverage the function approximation powers of Neural Networks to simplify the GMR formulation to a linear Mixture of Gaussian formulation. 
    
Through our experiments, we see that our approach successfully captures the best of recurrent, multi-modal, and reactive representations for learning short-horizon Human-Robot Interactions from demonstrations. 
% \end{enumerate}
We find that MoVEInt generates highly accurate robot behaviors without explicit action labels, which is more natural as humans also internally infer what our interaction partner is doing and adapt to it without explicitly communicating the action being done. We validate our predictive performance on a variety of physical HRI scenarios such as handshakes, fistbumps, and robot-to-human handovers. We further demonstrate the efficacy of MoVEInt on a real-world interaction scenario for bimanual (dual-arm) robot-to-human handovers.

% Organization paragraph
% The rest of the paper is organized as follows.

\section{Foundations}
\label{sec:Foundations}
% We first explain Variational Autoencoders (VAEs) (Section~\ref{ssec:vae}), which we use for learning the embeddings of the robot motions, followed by Mixture Density Networks (MDNs) (Section~\ref{ssec:MDN}), which are the key component in our policy representation.

\subsection{Variational Autoencoders}
\label{ssec:vae}

Variational Autoencoders (VAEs)~\cite{kingma2013auto,rezende2014stochastic} are a class of neural networks that learn to reconstruct inputs via latent representations in an unsupervised, probabilistic way. The inputs \enquote{$\boldsymbol{x}$} are encoded into lower dimensional latent embeddings \enquote{$\boldsymbol{z}$} that a decoder uses to reconstruct the input. A prior distribution, typically a standard normal distribution $p(\boldsymbol{z})=\mathcal{N}(\boldsymbol{z}; \boldsymbol{0, I})$, is enforced over the latent space during training. The goal is to estimate the true posterior $p(\boldsymbol{z|x})$ using a neural network $q(\boldsymbol{z|x})$, which is trained by minimizing the Kullback-Leibler (KL) divergence between them

\begin{equation}
    KL(q(\boldsymbol{z|x})||p(\boldsymbol{z|x})) = \mathbb{E}_q[\log\frac{ q(\boldsymbol{z|x})}{p(\boldsymbol{x,z})}] + \log p(\boldsymbol{x})
\end{equation}
which  can be re-written as 
\begin{equation}
    \label{eq:evidence_log}
    \log p(\boldsymbol{x}) = KL(q(\boldsymbol{z|x})||p(\boldsymbol{z|x})) + \mathbb{E}_q[\log\frac{p(\boldsymbol{x,z})}{q(\boldsymbol{z|x})}].
\end{equation}

The KL divergence is always non-negative; therefore, the second term in Eq.~\ref{eq:evidence_log} acts as a lower bound. Maximizing it would effectively maximize the log-likelihood of the data distribution or evidence and is hence called the Evidence Lower Bound (ELBO), which can be written as
\begin{equation}
    \label{eq:elbo}
    \mathbb{E}_q[\log\frac{p(\boldsymbol{x,z})}{q(\boldsymbol{z|x})}] = \mathbb{E}_q\log p(\boldsymbol{x|z}) - KL(q(\boldsymbol{z|x})||p(\boldsymbol{z}))
\end{equation}
The first term aims to reconstruct the input via samples decoded from the posterior. The second term is the KL divergence between the prior and the posterior, which regularizes the learning. %To prevent over-regularization, the KL divergence term is also weighted down with a factor $\beta$. 
Further information can be found in~\cite{kingma2013auto,rezende2014stochastic,higgins2016beta}. %Typically, $p(\boldsymbol{z})$ is represented via a standard normal distribution $\mathcal{N}(\mathbf{0}, \mathbf{I })$ with zero mean and unit variance.

\subsection{Mixture Density Networks}
\label{ssec:MDN}

Mixture Density Networks (MDNs)~\cite{bishop1994mixture} is a probabilistic neural network architecture that encapsulates the representation powers of neural networks with the modular advantages that come with Gaussian Mixture Models (GMMs). MDNs parameterize a typical supervised regression problem of predicting an output distribution $p(\boldsymbol{y}|\boldsymbol{x})~=~\mathcal{N}(\boldsymbol{y}|\boldsymbol{\mu}(\boldsymbol{x}), \boldsymbol{\sigma}^2(\boldsymbol{x})))$ as a multimodal distribution by predicting a set of means and variances $(\boldsymbol{\mu}_i(\boldsymbol{x}), \boldsymbol{\sigma}_i^2(\boldsymbol{x}))$ corresponding to the mixture distribution along with the mixture coefficients $\alpha_i(\boldsymbol{x})$ corresponding to the relative weights of the mixture distributions.

% \begin{figure}[h!]
%     \centering
%     \includegraphics[width=\linewidth]{img/MDN.png}
%     \caption{Representation of Mixture Density Networks~\cite{vossen2018probabilistic}}
%     \label{fig:MDN}
% \end{figure}

% While MDNs are trained in a discriminative manner by maximizing the log-likelihood of the data, they can be used in a generative manner, like a GMM, by first sampling from a categorical distribution parameterized by $\alpha_i$ and then sampling from the corresponding Mixture component

% \begin{equation}
% \label{eq:mdn-sampling}
%     \begin{gathered}
%         c \sim Cat(\alpha_i(\boldsymbol{x}))\\
%         \boldsymbol{y} \sim \mathcal{N}(\boldsymbol{\mu}_c(\boldsymbol{x}), \boldsymbol{\sigma}_c^2(\boldsymbol{x})).
%     \end{gathered}
% \end{equation}

\begin{figure*}[t!]
    \centering
    \includegraphics[width=0.9\linewidth]{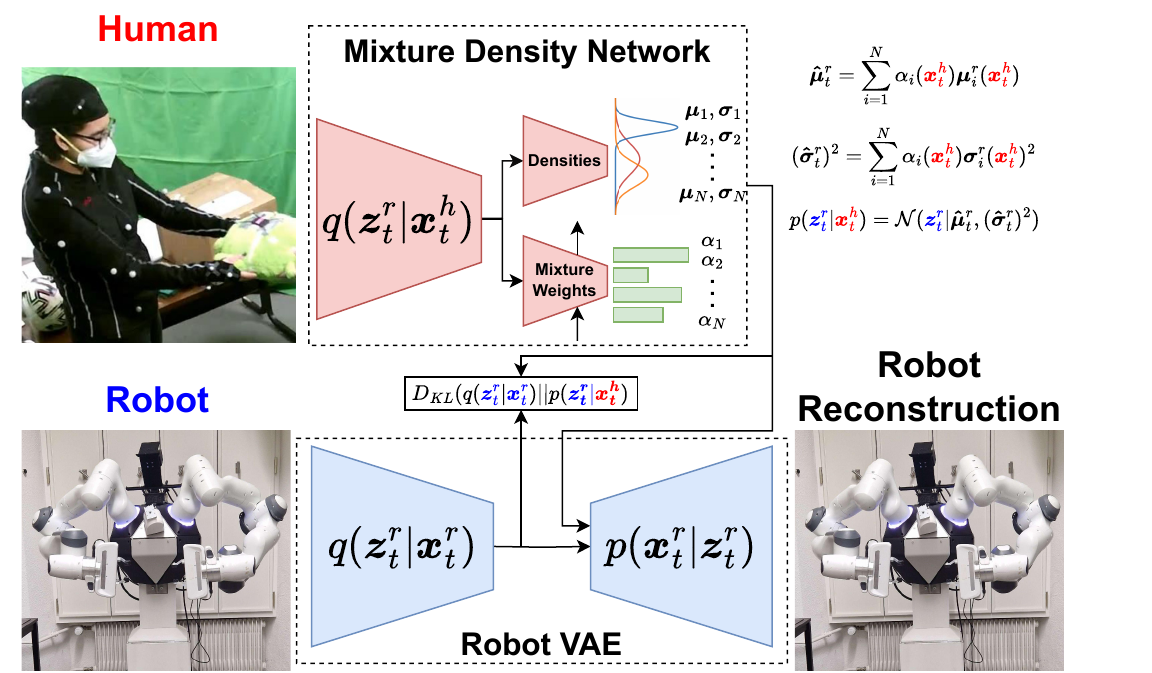}
    \caption{Overview of our approach \enquote{MoVEInt}. We train a reactive policy using a Mixture Density Network (MDN) to predict latent space robot actions from human observations. The MDN policy is used not just for reactively generating the robot's actions, but also to regularize a VAE that learns a latent representation of the robot's actions. This regularization ensures that the learned robot representation matches the predicted MDN policy and also ensures that the robot VAE learns to decode samples from the MDN policy.
    % The robot actions are learned via an Autoencoder which can then be executed on a robot, thereby enabling a well-coordinated, responsive interactive behavior.
    }
    \label{fig:overview}
    \vspace{-2em}
\end{figure*}

% \section{Recurrent Mixture Density VAEs for HRI}
\section{Mixture of Variational Experts for Learning Human-Robot Interactions from Demonstrations}
\label{sec:approach}
In this section, we present MoVEInt, a novel framework that learns latent space policies in a Mixture of Experts fashion for modeling the shared dynamics of a human and a robot in HRI tasks. This process can be seen in Fig.~\ref{fig:overview}. We aim to model the dynamics of HRI tasks via shared latent representations of a human and a robot in a way that captures the multimodality of the demonstrations and subsequently predicts the robot's motions in a reactive manner. To do so, we use an MDN that takes the human observations as input and predicts multiple latent policies, thereby enabling a multimodal output, and subsequently, the relative weights for each policy so that they can be effectively combined. For learning a shared latent representation between the human and the robot, we train a VAE over the robot motions and regularize the VAE with the predicted policy from the MDN, thereby learning the robot embeddings and the subsequent human-conditioned policy predictions in a cohesive manner.

We motivate the use of MDNs in Sec.~\ref{ssec:gmr-mdn} by showing the equivalence of MDNs with Gaussian Mixture Regression (GMR) for HRI. We then explain learning the robot motion embeddings (Sec.~\ref{ssec:robot-vae}), and then show how to train reactive policies for HRI (Sec.~\ref{ssec:policy-learning}).
% \subsection{Overview}
% \label{ssec:overview}
% We aim to model the joint latent space dynamics in HRI scenarios via a multimodal reactive policy. To do so, we train a Mixture Density Network that takes the human observations as input and predicts multiple latent policies, enabling a multimodal output, and subsequently, the relative weights for each policy so that they can be effectively combined. For learning a shared latent representation between the human and the robot, we train a VAE over the robot motions and regularize the VAE with the predicted policy from the MDN, thereby learning the robot embeddings and the subsequent human-conditioned policy predictions in a cohesive manner. This process can be seen in Fig.~\ref{fig:overview}. 
% For an easier understanding, 
We denote the human variables in \textcolor{red}{red} with the superscript \textcolor{red}{$h$} and the robot variables in \textcolor{blue}{blue} with the superscript \textcolor{blue}{$r$}.

\subsection{GMR-based Interaction Dynamics with MDNs}
\label{ssec:gmr-mdn}

% {\color{blue}Give a good connection to MILD, calinon and other GMM approaches to motivate why GMR is important}

Learning a joint distribution over the degrees of freedom of a human and a robot has been widely used in learning HRI from demonstrations~\cite{calinon2016tutorial,evrard2009teaching,calinon2009learning,maeda2014learning,ewerton2015learning,prasad2022mild,prasad2023learning}. With a joint distribution, GMR provides a mathematically sound formulation of predicting the conditional distribution of the robot actions. %While it can be used with just a single joint distribution, we explain the general case of how GMR is applied with a mixture model. 
When using a Mixture of $N$ Gaussian components $\{\boldsymbol{\mu}_i, \boldsymbol{\Sigma}_i\}$ that model a joint distribution of the Human and Robot trajectories, the distribution can be decomposed into the marginals for the human and the robot
\begin{equation}
\label{eq:gmr}
    \boldsymbol{\mu}_i = \begin{bmatrix}
\boldsymbol{\mu}_i^h\\
\boldsymbol{\mu}_i^r
\end{bmatrix}; \boldsymbol{\Sigma}_i = \begin{bmatrix}
\boldsymbol{\Sigma}^{hh}_i & \boldsymbol{\Sigma}^{hr}_i\\
\boldsymbol{\Sigma}^{rh}_i & \boldsymbol{\Sigma}^{rr}_i
\end{bmatrix}
\end{equation}
% where $h$ indicates the human and $r$ indicates the robot. 
In such a formulation, during test time, the robot motions can be generated reactively by conditioning the distribution using Gaussian Mixture Regression (GMR)~\cite{calinon2016tutorial,sung2004gaussian}\footnote{The covariance in the GMR Formulation in~\cite{calinon2016tutorial}, which is derived from~\cite{sung2004gaussian}, can be obtained by simplifying Eq.~\ref{eq:gmr-conditioning} as shown in the appendix.}

\begin{equation}
\label{eq:gmr-conditioning}
\begin{gathered}
% \begin{center}
    % \label{eq:gmr-conditioning-K-i}
    \boldsymbol{K}_i = \boldsymbol{\Sigma}^{rh}_i(\boldsymbol{\Sigma}^{hh}_i)^{-1}\\
    % \label{eq:gmr-conditioning-mu-i}
% \begin{split}
    \boldsymbol{\hat{\mu}}^r_i = \boldsymbol{\mu}^r_i + \boldsymbol{K}_i(\textcolor{red}{\boldsymbol{z}^h_t} - \boldsymbol{\mu}^h_i) \\%\hfill \forall i \in [1 \dots N]\\
    % \label{eq:gmr-conditioning-mu-t}
    \boldsymbol{\hat{\mu}}^r_t = \sum_{i=1}^N \alpha_i(\textcolor{red}{\boldsymbol{z}^h_t}) \boldsymbol{\hat{\mu}}^r_i\\
    % \label{eq:gmr-conditioning-sigma-i}
    \boldsymbol{\hat{\Sigma}}^r_i = \boldsymbol{\Sigma}^{rr}_i - \boldsymbol{K}_i\boldsymbol{\Sigma}^{hr}_i + (\boldsymbol{\hat{\mu}}^r_i - \boldsymbol{\hat{\mu}}^r_t)(\boldsymbol{\hat{\mu}}^r_i - \boldsymbol{\hat{\mu}}^r_t)^T\\%\hfill \forall i \in [1 \dots N]\\
    % \label{eq:gmr-conditioning-sigma-t}
    \boldsymbol{\hat{\Sigma}}^r_t = \sum_{i=1}^N \alpha_i(\textcolor{red}{\boldsymbol{z}^h_t}) \boldsymbol{\hat{\Sigma}}^r_i\\
    % \label{eq:gmr-conditioning-pz2z1}
    p(\textcolor{blue}{\boldsymbol{z}^r_t}|\textcolor{red}{\boldsymbol{z}^h_t}) = \mathcal{N}(\textcolor{blue}{\boldsymbol{z}^r_t} | \boldsymbol{\hat{\mu}}^r_t, \boldsymbol{\hat{\Sigma}}^r_t)
% \end{split}
% \end{center}
\end{gathered}
\end{equation}

where $\alpha_i(\textcolor{red}{\boldsymbol{z}^h_t})$ is the relative weight of each component. % calculated using the marginal distribution of the observed human agent. %When using GMMs, it represents the responsibilities of each Mixture Component $\alpha_i(\textcolor{red}{\boldsymbol{z}^h_t}) \propto \pi_i\mathcal{N}(\textcolor{red}{\boldsymbol{z}^h_t}|\boldsymbol{\mu}^h_i, \boldsymbol{\Sigma}_k^{hh})$. In the case of HMMs, $\alpha_i(\boldsymbol{z}^h_t)$ is the forward variable representing the belief state at the given timestep in a recurrent manner, given as $\alpha_i(\textcolor{red}{\boldsymbol{z}^h_t})~\propto~\mathcal{N}(\textcolor{red}{\boldsymbol{z}^h_t}|\boldsymbol{\mu}^h_i,\boldsymbol{\Sigma}_k^{h})\sum_{j=1}^N\mathcal{T}_{i,j}\alpha_j(\textcolor{red}{\boldsymbol{z}^h_{t-1}})$ where $\mathcal{T}_{i,j}$ denotes the transition probability of the system transitioning from state $i$ to $j$. 

Revisiting Eq.~\ref{eq:gmr-conditioning}, given the linear dependence of $\boldsymbol{\hat{\mu}}^r_i$ on $\textcolor{red}{\boldsymbol{z}^h_t}$, we can therefore consider $\boldsymbol{\hat{\mu}}^r_i$ as a direct output of a neural network. % $q_h^{\mu}(\textcolor{red}{\boldsymbol{x}^h_t})$.
% \begin{equation}
% \begin{aligned}
%     \boldsymbol{\hat{\mu}}^r_i &= \boldsymbol{\mu}^r_i + \boldsymbol{K}_i(\boldsymbol{z}^h_t - \boldsymbol{\mu}^h_i) \\
%     &= \underbrace{\boldsymbol{K}_i}_{\boldsymbol{W}_i} \boldsymbol{z}^h_t + \underbrace{(\boldsymbol{\mu}^r_i - \boldsymbol{K}_i \boldsymbol{\mu}^h_i)}_{\boldsymbol{b}_i}\\
%     & = \boldsymbol{W}_i \boldsymbol{z}^h_t + \boldsymbol{b}_i
% \end{aligned}
% \label{eq:mu-linear-layer}
% \end{equation}
Although the covariance matrix has a quadratic relationship with $\textcolor{red}{\boldsymbol{z}^h_t}$, considering Neural Networks are powerful function approximations, we assume that our network can adequately approximate a diagonalized form of the covariance matrices $\boldsymbol{\hat{\Sigma}}^r_i$~\cite{mclachlan1988mixture,zhou2020movement}. To consider the temporal aspect of learning such trajectories from demonstrations, rather than using the Mixture Model coefficients, $\alpha_i(\textcolor{red}{\boldsymbol{z}^h_t})$ can be approximated in a temporal manner using Hidden Markov Models (HMMs)~\cite{calinon2016tutorial} $\alpha_i(\textcolor{red}{\boldsymbol{z}^h_t}) = \mathcal{N}(\textcolor{red}{\boldsymbol{z}^h_t};\boldsymbol{\mu}_i, \boldsymbol{\Sigma}_i)\sum_{j=1}^N \alpha_j(\textcolor{red}{\boldsymbol{z}_{t-1}})\mathcal{T}_{j,i}$ where the parameters $(\boldsymbol{\mu}_i, \boldsymbol{\Sigma}_i, \mathcal{T}_{j,i})$ are the means and covariances of the underlying Gaussian states and the transition probabilities between the states respectively.

Given that there exist parallels between HMMs and RNNs~\cite{baucum2021hidden,hiraoka2021recurrent,chiu2020scaling} and to capture the recurrence in predicting $\alpha_i(\textcolor{red}{\boldsymbol{z}^h_t})$, we use a recurrent layer for predicting the mixing coefficients. Using the predictions of the mixture model parameters $(\boldsymbol{\mu}^r_i(\textcolor{red}{\boldsymbol{x}^h_t}), \boldsymbol{\sigma}^r_i(\textcolor{red}{\boldsymbol{x}^h_t})^2)$ and coefficients $\alpha_i(\textcolor{red}{\boldsymbol{x}^h_t})$ from the MDN \footnote{Since the relation between $\textcolor{red}{\boldsymbol{x}^h_t}$ and $\textcolor{red}{\boldsymbol{z}^h_t}$ is deterministic, for ease of notation, we show the prediction of the MDN components with $\textcolor{red}{\boldsymbol{x}^h_t}$}, we can re-write Eq.~\ref{eq:gmr-conditioning} as

\begin{equation}
\label{eq:gmr-mdn}
\begin{gathered}
    \boldsymbol{\hat{\mu}}^r_t = \sum_{i=1}^N \alpha_i(\textcolor{red}{\boldsymbol{x}^h_t}) \boldsymbol{\mu}^r_i(\textcolor{red}{\boldsymbol{x}^h_t})\\
    (\boldsymbol{\hat{\sigma}}^r_t)^2 = \sum_{i=1}^N \alpha_i(\textcolor{red}{\boldsymbol{x}^h_t}) \boldsymbol{\sigma}^r_i(\textcolor{red}{\boldsymbol{x}^h_t})^2\\
    p(\textcolor{blue}{\boldsymbol{z}^r_t}|\textcolor{red}{\boldsymbol{x}^h_t}) = \mathcal{N}(\textcolor{blue}{\boldsymbol{z}^r_t} | \boldsymbol{\hat{\mu}}^r_t, (\boldsymbol{\hat{\sigma}}^r_t)^2)
% \end{split}
% \end{center}
\end{gathered}
\end{equation}

which is then used as a prior for regularizing the VAE and training the decoder to reconstruct latent samples obtained after observing the human partner.

\subsection{Robot Motion Embeddings}
\label{ssec:robot-vae}
To learn a meaningful representation of the robot's actions, we train a VAE to reconstruct the robot's actions at each timestep. Typically, in VAEs, a standard normal distribution is used as the latent space prior $p(\boldsymbol{z}) = \mathcal{N}(\boldsymbol{0},\boldsymbol{I})$. Rather than forcing an uninformative standard normal prior as in Eq.~\ref{eq:elbo}, we use the reactive policy predicted from the human observations by the MDN (Eq.~\ref{eq:gmr-mdn}) to regularize the VAE's posterior $KL(q(\textcolor{blue}{\boldsymbol{z}^r_t}|\textcolor{blue}{\boldsymbol{x}^r_t})||p(\textcolor{blue}{\boldsymbol{z}^r_t}|\textcolor{red}{\boldsymbol{x}^h_t}))$, thereby learning a task-oriented latent space that is in line with the interaction dynamics. Our ELBO for training the robot VAE can be written as 

\begin{equation}
    \label{eq:robot-elbo}
    \begin{aligned}
    ELBO_t^r &= \mathbb{E}_{q(\textcolor{blue}{\boldsymbol{z}^r_t}|\textcolor{blue}{\boldsymbol{x}^r_t})}[\log p(\textcolor{blue}{\boldsymbol{x}^r_t}|\textcolor{blue}{\boldsymbol{z}^r_t})] \\
    &- \beta KL(q(\textcolor{blue}{\boldsymbol{z}^r_t}|\textcolor{blue}{\boldsymbol{x}^r_t})||p(\textcolor{blue}{\boldsymbol{z}^r_t}|\textcolor{red}{\boldsymbol{x}^h_t}))
    \end{aligned}
\end{equation}

Where $\beta$ is a relative weight used to ensure numerical stability between the KL divergence term and the image reconstruction term~\cite{higgins2016beta}.
% To ensure that the robot embeddings at a given timestep correspond correctly to the respective segment, we use the segment labels for calculating the prior in Eq.~\ref{eq:robot-elbo} rather than the segment predictions in Eq.~\ref{eq:latent-prior}.

\subsection{Reactive Motion Generation}
\label{ssec:policy-learning}

We aim to learn a policy for reactively generating the robot's latent trajectory based on human observations $p(\textcolor{blue}{\boldsymbol{z}^r_t}|\textcolor{red}{\boldsymbol{x}^h_t})$. We do so in a Behavior Cloning Paradigm by maximizing the probability of the observed trajectories w.r.t. the current policy $\mathcal{L}^{BC}_t = -\mathbb{E}_{\textcolor{blue}{\boldsymbol{z}^r_t}\sim p(\textcolor{blue}{\boldsymbol{z}^r_t}|\textcolor{red}{\boldsymbol{x}^h_t})} p(\textcolor{blue}{\boldsymbol{x}^r_t}|\textcolor{blue}{\boldsymbol{z}^r_t})$ wherein we first draw samples from the current policy $\textcolor{blue}{\boldsymbol{z}^r_t}\sim p(\textcolor{blue}{\boldsymbol{z}^r_t}|\textcolor{red}{\boldsymbol{x}^h_t})$ which we then reconstruct $p(\textcolor{blue}{\boldsymbol{x}^r_t}|\textcolor{blue}{\boldsymbol{z}^r_t})$, thereby enabling the decoder to reconstruct latent samples obtained after observing the human, as done during test time.

However, as highlighted in~\cite{zhou2020movement}, MDN policy representations are prone to mode collapses. Therefore, to ensure adequate separation between the modes so that we can learn a diverse range of actions, we employ a contrastive loss at each timestep. The contrastive loss pushes the means of each mixture component further away, while maintaining temporal similarity by pushing embeddings that are closer in time nearer to each other. Further, as done in~\cite{zhou2020movement}, we add entropy cost to ensure a balanced prediction of the mixture coefficients. Our separation loss can be written as 
\begin{equation}
\begin{aligned}
    \mathcal{L}^{sep}_t &= \underbrace{\sum_{i=1}^{N-1}\sum_{j=i+1}^N exp(-\lVert\boldsymbol{\mu}^r_i(\textcolor{red}{\boldsymbol{x}^h_t}) - \boldsymbol{\mu}^r_j(\textcolor{red}{\boldsymbol{x}^h_t})\rVert^2)}_{\text{separation of means}} \\
    & + \underbrace{1 - \frac{1}{N}\sum_{i=1}^N exp(-\lVert\boldsymbol{\mu}^r_i(\textcolor{red}{\boldsymbol{x}^h_t}) - \boldsymbol{\mu}^r_i(\textcolor{red}{\boldsymbol{x}^h_{t-1}})\rVert^2)}_{\text{temporal closeness}} \\
    &+ \underbrace{\sum_{i=1}^N \alpha_i(\textcolor{red}{\boldsymbol{x}^h_t}) \ln \alpha_i(\textcolor{red}{\boldsymbol{x}^h_t})}_{\textit{entropy cost}}
\end{aligned}
\end{equation}

Our final loss consists of the Behavior Cloning loss, the ELBO of the robot VAE, and the separation loss

\begin{equation}
\label{eq:final-loss}
    \sum_{t=1}^T \left[\mathcal{L}^{BC}_t - ELBO_t^r + \beta\mathcal{L}^{sep}_t \right]
\end{equation}

where $\beta$ is the same KL weight factor used in Eq.~\ref{eq:robot-elbo}. Our overall training procedure is shown in Alg.~\ref{alg:training-algo}.

\begin{algorithm}[h!]
\caption{Learning a Reactive Latent Policy for Human-Robot Interaction}
\label{alg:training-algo}
\small
    \KwData{A set of human and robot trajectories $\boldsymbol{X} =\{\boldsymbol{X}^h, \boldsymbol{X}^r\}$}
    \KwResult{MDN Policy Network, Robot VAE}
    
    \While{not converged}{
        \For {$\boldsymbol{x}^h_{1:T}, \boldsymbol{x}^r_{1:T} \in \boldsymbol{X}$}{
            Compute the MDN policy $p(\textcolor{blue}{\boldsymbol{z}^r_t}|\textcolor{red}{\boldsymbol{x}^h_t})$ (Eq.~\ref{eq:gmr-mdn}) \\
            Compute the robot VAE posterior $p(\textcolor{blue}{\boldsymbol{z}^r_t}|\textcolor{blue}{\boldsymbol{x}^r_t})$ (Eq.~\ref{eq:robot-elbo}) \\
            Reconstruct samples from $p(\textcolor{blue}{\boldsymbol{z}^r_t}|\textcolor{red}{\boldsymbol{x}^h_t})$ and $p(\textcolor{blue}{\boldsymbol{z}^r_t}|\textcolor{blue}{\boldsymbol{x}^r_t})$\\
            Minimize the loss in Eq.~\ref{eq:final-loss} to update the network 
        }
    }
\end{algorithm}

During test time, given human observations $\textcolor{red}{\boldsymbol{x}^h_t}$, we compute the latent policy from the MDN $\textcolor{blue}{\boldsymbol{z}^r_t}~=~p(\textcolor{blue}{\boldsymbol{z}^r_t}|\textcolor{red}{\boldsymbol{x}^h_t})$ which is then decoded to obtain the robot action $p(\textcolor{blue}{\boldsymbol{x}^r_t}|\textcolor{blue}{\boldsymbol{z}^r_t})$.

\section{Experiments and Results}
\label{sec:exp}
In this section, we first explain the datasets we use (Sec.~\ref{sec:dataset}), then we highlight the implementation details (Sec.~\ref{sec:training-details}), %the robot setup for showcasing our approach (Sec.~\ref{sec:robot-setup}), 
and finally present our results (Sec.~\ref{sec:results}).

\subsection{Dataset}
\label{sec:dataset}

\subsubsection{B\"utepage et al.}~\cite{butepage2020imitating}
%\footnote{\url{https://github.com/jbutepage/human_robot_interaction_data}}
\label{ssec:buetepage-dataset}record HHI and HRI demonstrations of 4 interactions: Waving, Handshaking, and two kinds of fist bumps. The first fistbump, called \enquote{Rocket Fistbump}, involves bumping fists at a low level and then raising them upwards while maintaining contact with each other. The second is called \enquote{Parachute Fistbump} in which partners bump their fists at a high level and bring them down while simultaneously oscillating the hands sideways, while in contact with each other. They additionally record demonstrations of these actions with a human partner interacting with an ABB YuMi-IRB 14000 robot controlled via kinesthetic teaching. We call this scenario as \enquote{HRI-Yumi}. %In the HHI scenario (and the HRI-Pepper scenario), there are 181 trajectories (32 - Waving, 38 - Handshake, 70 - Rocket Fistbump, 49 - Parachute Fistbump) of which 80\% of the trajectories (149 trajectories) are used for training and the rest (32 trajectories) for testing. In the HRI-Yumi scenario, there are 41 trajectories (10 each for Waving, Handshaking, and Rocket Fistbump and 11 for Parachute Fistbump) of which we use a similar split with 32 trajectories for training and 9 for testing. 

We further adapt the HHI demonstrations for the humanoid robot Pepper~\cite{pandey2018mass} by extracting the joint angles from one of the human partner's skeletons from the HHI data for the Pepper robot using the similarities in DoFs between a human and Pepper~\cite{fritsche2015first,prasad2021learning}, which we denote this as \enquote{HRI-Pepper}. Given the smaller number of trajectories in the HRI-Yumi scenario, we initialize the network with the weights from the model trained for the HHI scenario.

\subsubsection{Nuitrack Skeleton Interaction Dataset}
\label{ssec:nuisi-dataset}(NuiSI)~\cite{prasad2023learning} is a dataset that we collected ourselves of the same 4 interactions as in~\cite{butepage2020imitating}. The skeleton data of two human partners interacting with one another is recorded using two Intel Realsense D435 cameras (one recording each human partner). We use Nuitrack~\cite{nuitrack} for tracking the upper body skeleton joints in each frame at 30Hz. As done with~\cite{butepage2020imitating}, we show results on both the HHI data as well as on the joint angles extracted for the Pepper robot from the skeleton trajectories for an HRI-Pepper scenario. 
%The data is first inspected manually to remove any trajectories where the tracking deteriorates. Finally, we have 44 trajectories (12 - Waving, 11 - Handshaking, 12 - Rocket Fistbump, 9 - Parachute Fistbump) of which we similarly use 80\% of the trajectories for training (33 trajectories) and the rest (11 trajectories) for testing. 

% \subsubsection{Human-Human Object Handovers Dataset}~\cite{kshirsagar2023dataset} is a dataset consisting of 12 pairs of participants performing object handovers with various objects with each partner taking the role of the giver and the receiver, leading to 24 pairs of handover partners. The dataset consists of both Unimanual (single-arm) and Bimanual (dual-arm) handovers tracked using motion capture. For the bimanual case, we use the original frequency of 120Hz. For the Unimanual case, we downsample the data to 20Hz as above. We explore a robot-to-human handover scenario and use the giver's observations corresponding to the robot actions and the receiver's observations corresponding to the human. %We remove trajectories that have occlusions in the tracking and use the last two pairs of participants for the testing set, leading to a total of 105 training trajectories and 12 testing trajectories.
\subsubsection{Human-Human Object Handovers Dataset}~\cite{kshirsagar2023dataset} is a dataset consisting of 12 pairs of participants performing object handovers with various objects, with each partner taking the role of the giver and the receiver, leading to 24 pairs of handover partners. The dataset consists of both Unimanual (single-arm) and Bimanual (dual-arm) handovers tracked using motion capture at 120Hz, which we downsample to 30Hz. We explore a Robot-to-Human handover scenario and use the giver's observations corresponding to the robot and the receiver's observations corresponding to the human. %We remove trajectories that have occlusions in the tracking and use the last two pairs of participants for the testing set.%, leading to a total of 105 training trajectories and 12 testing trajectories from the bimanual data and 63 training trajectories and 12 testing trajectories from the unimanual data.

\begin{table}[h!]
    \centering
    \begin{tabular}{|c|c|c|c|}
    \hline
         \multirow{2}{*}{Dataset} & Downsampled & \multicolumn{2}{c|}{No. of Trajectories} \\ \cline{3-4}
 & FPS (Hz) & Training & Testing\\ \hline
         HHI, HRI-Pepper~\cite{butepage2020imitating} & 20 & 149 & 32 \\ \hline
         HRI-Yumi~\cite{butepage2020imitating} & 20 & 32 & 9 \\ \hline
         HHI, HRI-Pepper (NuiSI~\cite{prasad2023learning}) & 30 & 33 & 11 \\ \hline
         HHI-Handovers~\cite{kshirsagar2023dataset} & 30 & 168 & 24 \\ \hline
    \end{tabular}
    \caption{Statistics of the different datasets used.}
    \label{tab:datasets}
    \vspace{-3em}
\end{table}

\input{mse_table}

\subsection{Implementation Details}
\label{sec:training-details}
After downsampling the data to the frequencies mentioned in Table~\ref{tab:datasets}, we use a time window of observations as the input for a given timestep, similar to~\cite{butepage2020imitating}. We use the 3D positions and the velocities (represented as position deltas) of the right arm joints (shoulder, elbow, and wrist), with the origin at the shoulder, leading to an input size of 90 dimensions (5x3x6: 5 timesteps, 3 joints, 6 dimensions) for a human partner. For the HHI-Handover scenario, we use both the left and right arm data (180 dimensions). For the HRI-Pepper and HRI-Yumi scenarios, we use a similar window of joint angles, leading to an input size of 20 dimensions (5x4) for the 4 joint angles of Pepper's right arm and an input size of 35 dimensions (5x7) for the 7 joint angles of Yumi's right arm. The reconstruction loss for the VAE and Behavior Cloning are calculated by decoding samples drawn from the respective distributions in a Monte Carlo fashion. %Sampling from the VAE posterior is straightforward using the re-parametrization trick~\cite{kingma2013auto}. For sampling from the MDN, we first need to draw a categorical sample, followed by drawing from the corresponding Gaussian (Eq.~\ref{eq:mdn-sampling}). To ensure the propagation of gradients when sampling from the categorical distribution, we use a Gumbel-Rao sampling mechanism~\cite{paulus2020rao} that is a variance reduction technique to the Straight-Through version of the Gumbel-Softmax estimator~\cite{jang2016categorical} via Rao-Blackwellisation. We use 100 samples for the Rao-Blackwellisation with the temperature set to 0.01 for scaling the Gumbel samples. 

For the dataset from~\cite{butepage2020imitating} and the NuiSI dataset~\cite{prasad2023learning}, we use a VAE with 2 hidden layers each in the encoder and decoder with a dimensionality of (40, 20) and (20, 40) respectively, with Leaky ReLU activations and a 5D latent space. For the VAE posterior and the MDN outputs, we predict the mean and the log of the standard deviation (logstd). We add a regularization of $10^{-6}$ to the standard deviation. The MDN has a similar structure as the VAE encoder but predicts 3 different means and logstds. For the mixture coefficients $\alpha_i$, to enable a recurrent nature of the predictions, the output of the MDN encoder is passed to a single-layer Gated Recurrent Unit (GRU) whose outputs are then passed through a linear layer followed by a softmax layer. For the NuiSI dataset, we initialize the model with the pre-trained weights trained on the dataset in~\cite{butepage2020imitating}.
% \begin{figure}[h!]
%     \centering
%     \includegraphics[width=0.95\linewidth]{img/network-sizes.pdf}
%     \caption{Network Architecture for training on the datasets~\cite{butepage2020imitating} and NuiSI. The Recurrent MDN for the human inputs is in red and the Robot VAE is in blue. (N - no. of components, Z - latent dimension).}
%     \label{fig:network}
% \end{figure}
% The network architechure for the dataset in~\cite{butepage2020imitating} and the NuiSI dataset is shown in Fig.~\ref{fig:network} where we use 3 mixture components in the MDN and a 5D latent space. 
For the Handover dataset~\cite{kshirsagar2023dataset}, we use double the number of states in each layer, namely 80 and 40 hidden states and a 10D latent space.

\subsection{Reactive Motion Generation Results}
\label{sec:results}

To show the advantage of our approach,  we compare MoVEInt with B\"utepage et al.~\cite{butepage2020imitating}, who use an LSTM-based approach as a latent regularization for reactive motion generation, which is close to a unimodal version of our approach. Further, we compare MoVEInt to MILD~\cite{prasad2023learning} which uses HMMs to capture the multimodal latent dynamics of interactive tasks. The efficacy of MoVEInt can be seen via the low mean squared error of the predicted robot motions (Table~\ref{tab:pred_mse}). 

We perform better than both MILD~\cite{prasad2023learning} and B\"utepage et al.~\cite{butepage2020imitating} on almost all interaction scenarios. We additionally want to highlight that on the HRI-Pepper scenarios, unlike MILD~\cite{prasad2023learning} and B\"utepage et al.~\cite{butepage2020imitating} where the pre-trained model from the HHI scenario is used, we train our model completely from scratch and still achieve better performance. Moreover, it is worth noting that both~\cite{butepage2020imitating} and MILD are trained in a partially supervised manner using the interaction labels. In~\cite{butepage2020imitating}, a one-hot label denoting the interaction being performed is given as an input to the network for generalizing to different interactions, whereas in MILD, a separate HMM is trained for each interaction. In contrast, MoVEInt is trained on all the tasks in a given dataset without any labels in a purely unsupervised manner and still achieves competitive results on the different datasets. 

\begin{figure*}[h!]
    \centering
    \begin{subfigure}[b]{0.5\textwidth}
         \centering
         \includegraphics[width=0.49\textwidth]{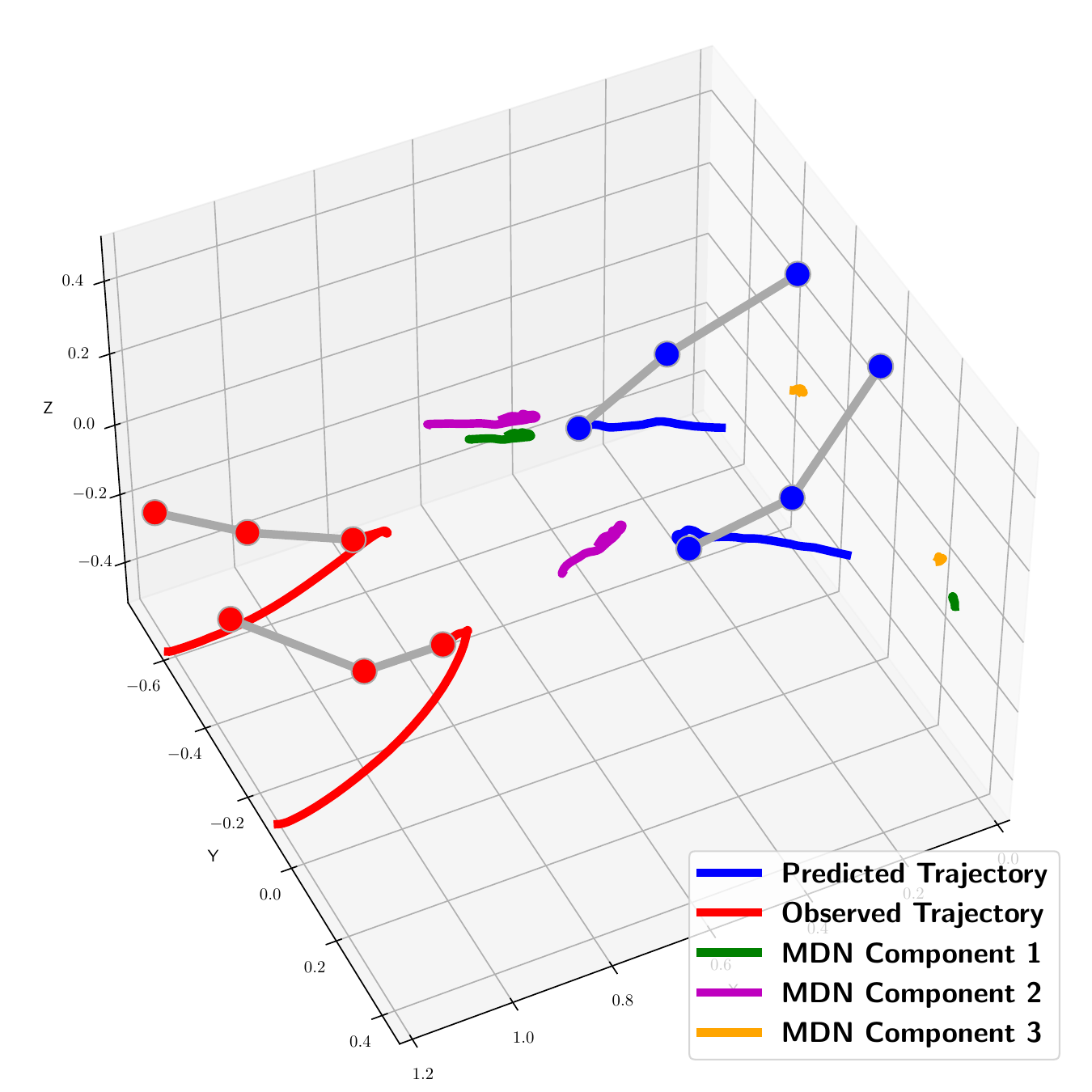}
         \includegraphics[width=0.49\textwidth]{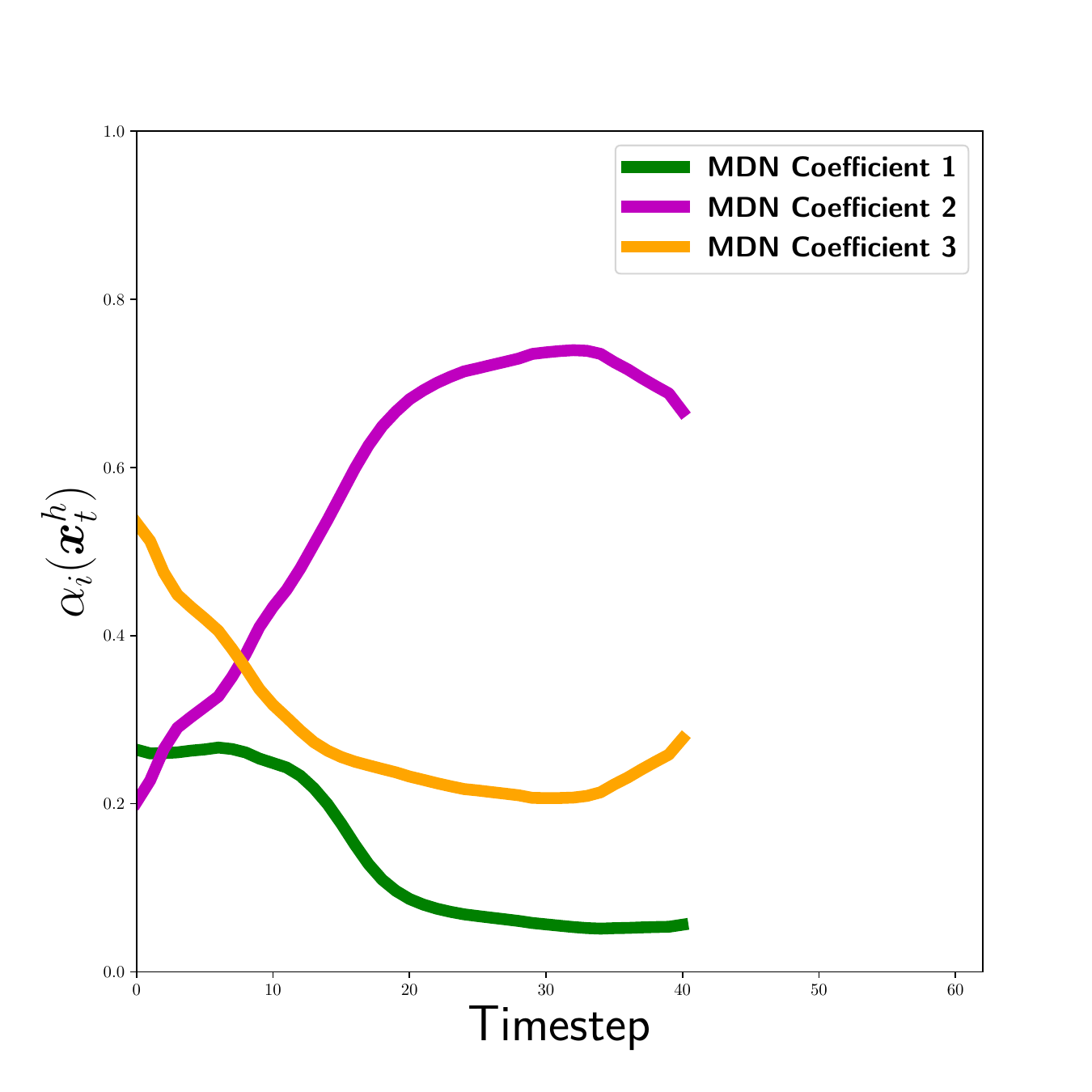}
         \caption{Example of a generated Bimanual Handovers}% generated from the dataset in~\cite{kshirsagar2023dataset}.}
         \label{fig:bimanual-hhi}
     \end{subfigure}\hfill
     \begin{subfigure}[b]{0.5\textwidth}
         \centering
         \includegraphics[width=0.49\textwidth]{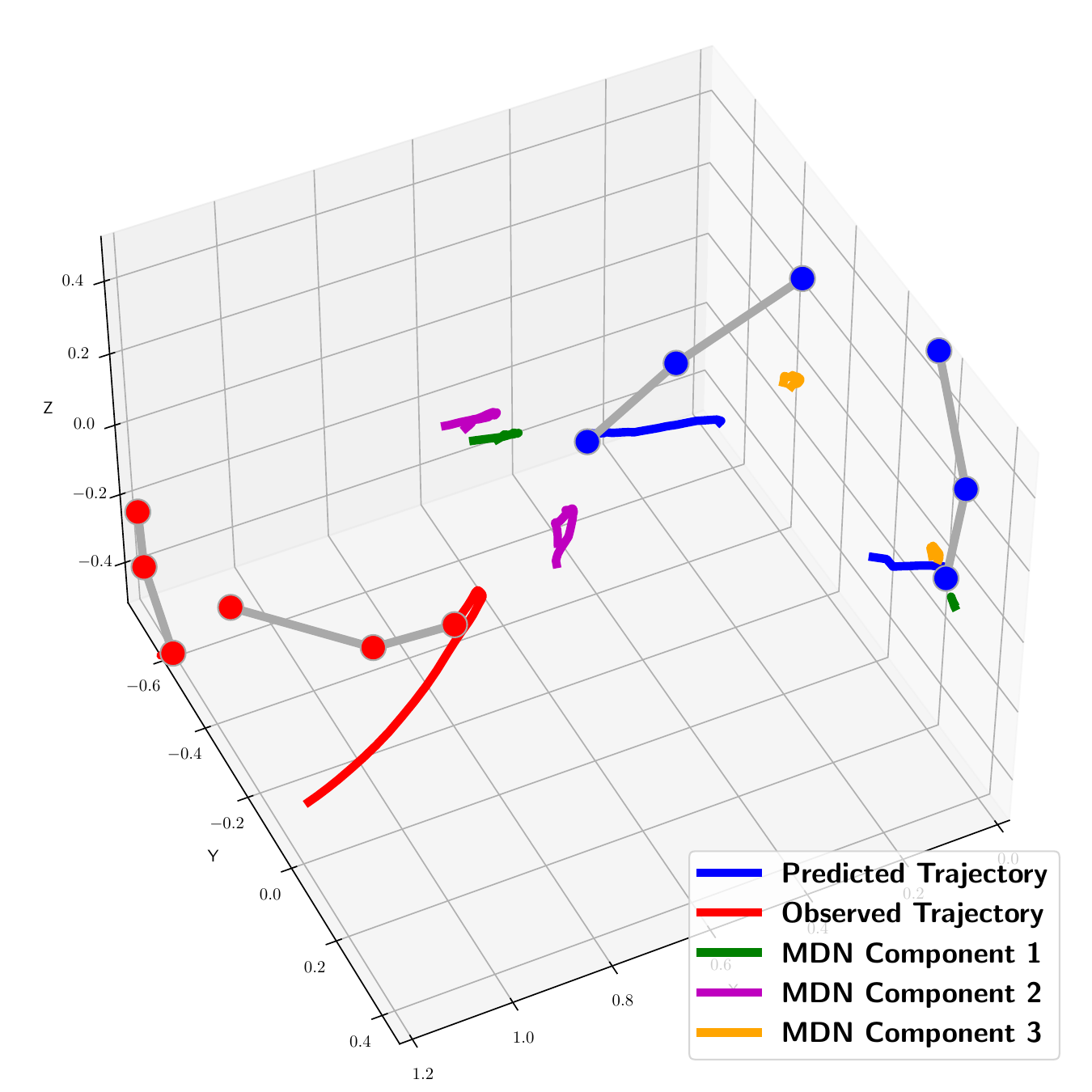}
         \includegraphics[width=0.49\textwidth]{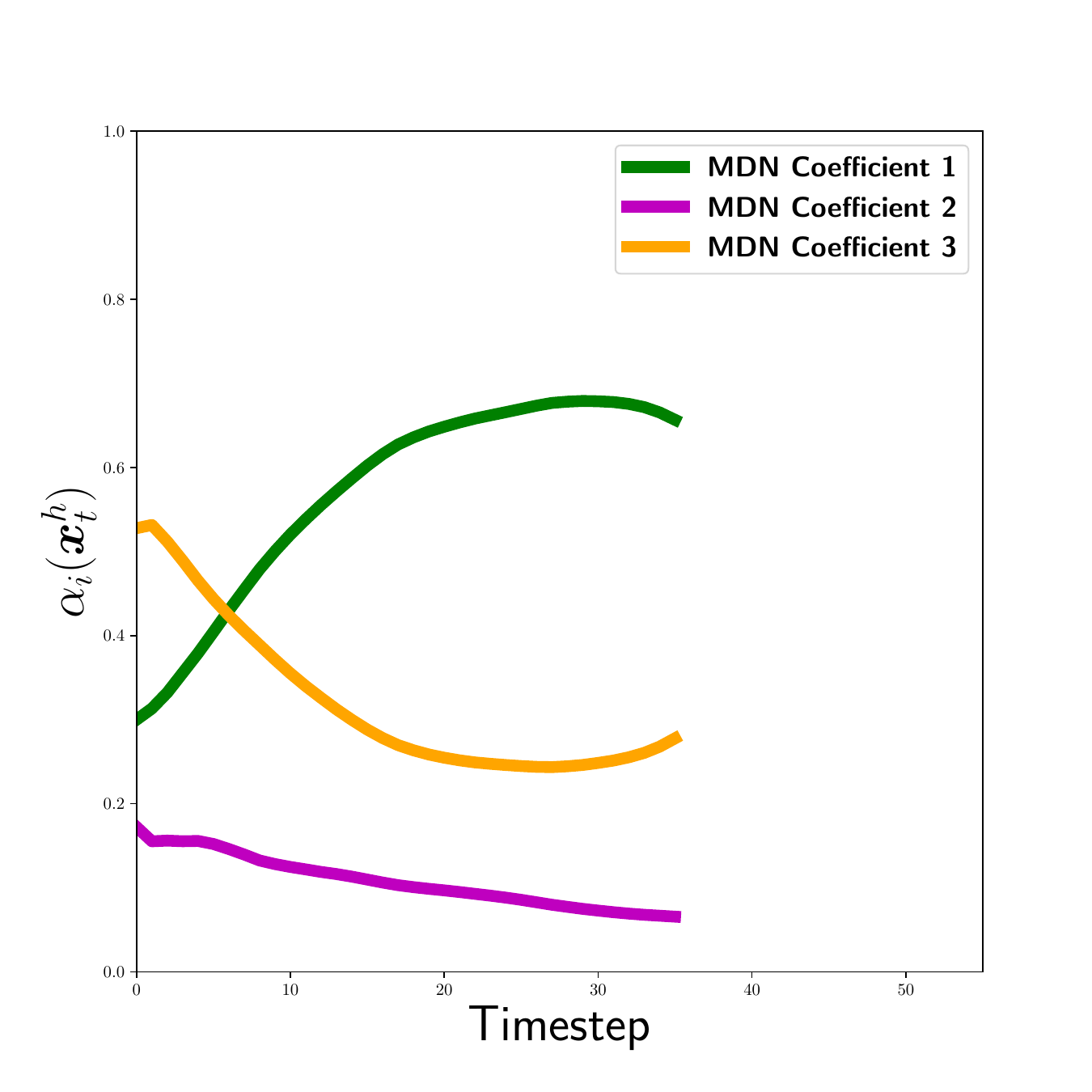}
         \caption{Example of a generated Unimanual Handovers}% generated from the dataset in~\cite{kshirsagar2023dataset}.}
         \label{fig:unimanual-hhi}
     \end{subfigure}
\caption{Sample trajectories generated by MoVEInt for the Bimanual and Unimanual Handovers in the HHI-Handovers dataset in~\cite{kshirsagar2023dataset}. The 3D plots show the reconstructed trajectories and the 2D plots show the corresponding progression of $\alpha_i(\textcolor{red}{\boldsymbol{x}^h_t})$ for the different components of the MDN. In the 3D plots, the observed trajectory of the receiver is shown in \textcolor{red}{red} and the generated trajectory of the giver is shown in \textcolor{blue}{blue} and the giver's corresponding ground truth is shown in black. The reconstruction of the individual latent components of the MDN are shown in \textcolor{mygreen}{green}, \textcolor{mymagenta}{magenta}, and \textcolor{myorange}{orange}. It can be seen that the learned components correspond to different parts of the task space. For example, \textcolor{mygreen}{green} denotes the hand locations for a unimanual handover, \textcolor{mymagenta}{magenta} denotes the hand locations for a bimanual handover, and \textcolor{myorange}{orange} denotes the static hand locations for the starting and ending neutral poses. In the 2D plot, it can be seen how the coefficients for components corresponding to bimanual (\textcolor{mymagenta}{magenta}) and unimanual (\textcolor{mygreen}{green}) get activated based on the interaction being performed, while the component corresponding to a neutral pose (\textcolor{myorange}{orange}) gets activated at the beginning of the interaction while both partners are static.}
    \label{fig:moveint-hhi}
    \vspace{-1em}
\end{figure*}

We additionally show some quantitative results of MoVEInt. We train a Handover model with just the hand trajectories whose predictions are used for reactive motion generation on a Bimanual Franka Emika Panda robot setup, \enquote{Kobo}, as shown in Fig.~\ref{fig:moveint-hri}. 
% We further demonstrate the Rocket-fistbump interaction in the HRI-Pepper scenario from the NuiSI dataset on the Pepper robot, which can be seen in Fig.~\ref{fig:pepper-fistbump}. 
Some additional examples of the trajectories generated by MoVEInt for bimanual and unimanual handovers from the HHI-Handovers dataset~\cite{kshirsagar2023dataset} are shown in Fig.~\ref{fig:bimanual-hhi} and ~\ref{fig:unimanual-hhi} respectively. Since MoVEInt is trained on all the interactions in a corresponding dataset, which, when coupled with the separation loss, learns a diverse and widespread set of components that cover the various demonstrations, as can be seen by the reconstruction of the individual components. Combining the components in the latent space subsequently leads to an accurate and suitable motion generated reactively (shown in blue). 
For further examples, please refer to our supplementary video.

\input{hri-kobo}

\subsection{User study}
To understand the effectiveness of our approach in the real world, we perform a feasibility study as a proof-of-concept with five users who perform bimanual robot-to-human handovers with the Kobo robot. We evaluate the ability of our approach to successfully generate a handover motion with three different objects where each participant interacts with the robot five times for each object (a total of 15 runs per participant). To maintain the object-centric nature of the interaction, we use a controller that tracks the mid-point of both end-effectors, thereby resembling tracking the object’s trajectory. This study was approved by the Ethics Commission at TU Darmstadt (Application EK 20/2023). For the study, we modify the Handovers dataset~\cite{kshirsagar2023dataset} such that the giver's trajectories fall within the task space limits of the Kobo robot. We train a separate model to use only the receiver's hand trajectories and subsequently predict the end-effector trajectory for the robot. This is given to an Object-centric Inverse Kinematics controller that tracks the motion of the mid-point between the end-effectors.

As shown in Table~\ref{tab:user_study}, our approach can generate successful handover trajectories for different users and different objects. We observed some failure cases due to sudden jumps in the predicted robot motions resulting from inaccuracies in perceiving the human, which would overshoot the robot's dynamic limits. However, this failure could be avoided by incorporating additional filters over the input and output data to MoVEInt. Some failures occurred because the object did not reach the exact vicinity of the human's hand location. This failure could be avoided by incorporating object-related information such as the size or weight, allowing the robot to gauge better when the handover is executed. Sometimes, the robot would retreat before the human could grasp the object if sufficient time had passed. One reason for this hasty retreating behavior could be that the recurrent network's hidden inputs overpower the observational input, causing the robot to follow the general motion of the handover seen during training and retreat. Such a failure could be mitigated by incorporating the robot state as part of the input.

\begin{table}[h!]
    \centering
    \begin{tabular}{|c|c|c|c|c|c|} \hline
        \multirow{2}{*}{\backslashbox{Object}{User ID}} & \multirow{2}{*}{\#1} & \multirow{2}{*}{\#2} & \multirow{2}{*}{\#3} & \multirow{2}{*}{\#4} &  Total \\
         &  &  &  &  & (per object)\\ \hline
        Stool & 5 & 4 & 5 & 5 & 19/20 \\ \hline
        Box & 4 & 5 & 4 & 4 & 17/20 \\ \hline
        Bedsheet & 4 & 5 & 3 & 3 & 15/20 \\ \hline 
        Total & \multirow{2}{*}{13/15} & \multirow{2}{*}{14/15} & \multirow{2}{*}{12/15} & \multirow{2}{*}{12/15} & \multirow{2}{*}{\textbf{51/60}} \\ 
        (per user) &  &  &  &  & \\ \hline
    \end{tabular}
    \caption{Number of successful handovers of each object by the Kobo robot to each user (total of 5 per object per user i.e. total of 20 per object and 15 per user).}
    \label{tab:user_study}
    \vspace{-2em}
\end{table}

% \begin{table}[h!]
%     \centering
%     \begin{tabular}{|c|c|c|c|c|c|} \hline
%         \multirow{2}{*}{\backslashbox{Object}{User ID}} & \multirow{2}{*}{1} & \multirow{2}{*}{2} & \multirow{2}{*}{3} & \multirow{2}{*}{4} &  Total \\
%          &  &  &  &  & (per object)\\ \hline
%         Box & 5 & 4 & 5 &  & \\ \hline
%         Stool & 4 & 5 & 4 &  & \\ \hline
%         Bedsheet & 4 & 5 & 3 &  & \\ \hline
%         Total (per user) & 13 & 14 & 12 &  & \\ \hline
%     \end{tabular}
%     \caption{Number of Successful handovers of each object by the Kobo robot to each user (total of 5 per object per user i.e. total of 20 per object and 15 per user).}
%     \label{tab:user_study}
%     \vspace{-2em}
% \end{table}

% completion/success rates
% reaction times.

\section{Conclusion and Future Work}
In this work, we presented \enquote{MoVEInt}, a novel deep generative Imitation Learning approach for learning Human-Robot Interaction from demonstrations in a Mixture of Experts fashion. We demonstrated 
% how the typically used GMR-based formulation of the interaction dynamics can yield itself to fall into 
the use of Mixture Density Networks (MDNs) as a multimodal policy representation in a shared latent space of the human and the robot. We showed how MoVEInt stems from the GMR-based formulation of predicting interaction dynamics used in HMM-based approaches to learning HRI. We showed how our MDN policy can predict multiple underlying policies and combine them to effectively generate response motions for the robot. We verified the efficacy of MoVEInt across a variety of interactive tasks, where we found that MoVEInt mostly outperformed other baselines that either use explicitly modular representations like an HMM or simple recurrent policy representations. Our experimental evaluation showcases the versatility of MoVEInt, which effectively combines explicitly modular distributions with recurrent policy representations for learning interaction dynamics.

The main focus of this article is to explore the use of MDNs as a latent policy representation for simplistic short-horizon interactions like handshakes, handovers, etc., and show the feasibility of our approach in learning various interactive behaviors. One drawback is that we currently do not explicitly study how robust the approach is to unknown behaviors of the human, or how it performs on more complicated tasks. 

Our next steps would be exploring how our approach scales to longer horizon tasks such as bi-directional handovers, collaborative sequential manipulation, or proactive tasks where the robot leads the interaction. This would be validated via a broader user study which would show the efficacy of our approach. Some extensions for extrapolating to such tasks can be explored by using better Generative models~\cite{wang2022co,sengadu2023dec,ng2023diffusion} or adding explicit constraints for combining reactive motion generation and planning~\cite{hansel2023hierarchical}. Our future work would also explore incorporating task-related constraints such as object information for handover grasps~\cite{christen2023learning,rosenberger2020object}, force information for enabling natural interactive behaviors~\cite{medina2016human,khanna2022human,bolotnikoval2018compliant}, 
and ergonomic and safety constraints for a more user-friendly interaction~\cite{lagomarsino2023maximising,rubagotti2022perceived}.

% Proper User study, object detection for human-to-robot handovers, bidirectional

% Incorporating MPC to ensure safe HRI 

% \section*{Acknowledgments}
% This work was supported by the German Research Foundation (DFG) Emmy Noether Programme (CH 2676/1-1), the German Federal Ministry of Education and Research (BMBF) Projects \enquote{IKIDA} (Grant no.: 01IS20045) and \enquote{KompAKI} (Grant no.: 02L19C150), the Excellence Program, \enquote{The Adaptive Mind}, of the Hessian Ministry of Higher Education, Science, Research and Art, the EU’s Horizon Europe projects \enquote{MANiBOT} (Grant no.: 101120823) and \enquote{ARISE} (Grant no.: 101135959).

% \newpage
\bibliographystyle{IEEEtran}
\bibliography{root}

\appendix
\section*{Gaussian Mixture Regression Simplification}

Here, we simplify the covariance calculation in the GMR formulation (Eq.~\ref{eq:gmr-conditioning}) to its equivalence to~\cite{calinon2016tutorial}.% The covariance of each Gaussian Mixture component $\boldsymbol{\hat{\Sigma}}^r_i$ can be expanded as

\begin{equation}
\centering
\begin{aligned}
    % \label{eq:gmr-conditioning-sigma-i}
    \boldsymbol{\hat{\Sigma}}^r_i = \boldsymbol{\Sigma}^{rr}_i - \boldsymbol{K}_i\boldsymbol{\Sigma}^{hr}_i &+ (\boldsymbol{\hat{\mu}}^r_i - \boldsymbol{\hat{\mu}}^r_t)(\boldsymbol{\hat{\mu}}^r_i - \boldsymbol{\hat{\mu}}^r_t)^T\\%\hfill \forall i \in [1 \dots N]\\
    = \boldsymbol{\Sigma}^{rr}_i - \boldsymbol{K}_i\boldsymbol{\Sigma}^{hr}_i & + \boldsymbol{\hat{\mu}}^r_i (\boldsymbol{\hat{\mu}}^r_i)^T 
    - \boldsymbol{\hat{\mu}}^r_i (\boldsymbol{\hat{\mu}}^r_t)^T \\
    &- \boldsymbol{\hat{\mu}}^r_t(\boldsymbol{\hat{\mu}}^r_i)^T +  \boldsymbol{\hat{\mu}}^r_t(\boldsymbol{\hat{\mu}}^r_t)^T\\
    \boldsymbol{\hat{\Sigma}}^r_t = \sum_{i=1}^N \alpha_i(\textcolor{red}{\boldsymbol{z}^h_t}) \boldsymbol{\hat{\Sigma}}^r_i\\
    = \sum_{i=1}^N \alpha_i(\textcolor{red}{\boldsymbol{z}^h_t})(\boldsymbol{\Sigma}^{rr}_i - \boldsymbol{K}&_i\boldsymbol{\Sigma}^{hr}_i) \\+ \sum_{i=1}^N \alpha_i(\textcolor{red}{\boldsymbol{z}^h_t})\boldsymbol{\hat{\mu}}^r_i (\boldsymbol{\hat{\mu}}^r_i)^T& 
    - (\sum_{i=1}^N \alpha_i(\textcolor{red}{\boldsymbol{z}^h_t})\boldsymbol{\hat{\mu}}^r_i) (\boldsymbol{\hat{\mu}}^r_t)^T \\
    - \boldsymbol{\hat{\mu}}^r_t\sum_{i=1}^N \alpha_i(\textcolor{red}{\boldsymbol{z}^h_t})(\boldsymbol{\hat{\mu}}^r_i)^T& + \boldsymbol{\hat{\mu}}^r_t(\boldsymbol{\hat{\mu}}^r_t)^T\sum_{i=1}^N \alpha_i(\textcolor{red}{\boldsymbol{z}^h_t})\\
    = \sum_{i=1}^N \alpha_i(\textcolor{red}{\boldsymbol{z}^h_t})(\boldsymbol{\Sigma}^{rr}_i - \boldsymbol{K}&_i\boldsymbol{\Sigma}^{hr}_i + \boldsymbol{\hat{\mu}}^r_i (\boldsymbol{\hat{\mu}}^r_i)^T) - \boldsymbol{\hat{\mu}}^r_t(\boldsymbol{\hat{\mu}}^r_t)^T
% \end{split}
% \end{center}
\end{aligned}
\end{equation}

since $\sum_{i=1}^N \alpha_i(\textcolor{red}{\boldsymbol{z}^h_t})(\boldsymbol{\hat{\mu}}^r_i) = \boldsymbol{\hat{\mu}}^r_t$ and $\sum_{i=1}^N \alpha_i(\textcolor{red}{\boldsymbol{z}^h_t}) = 1$, therefore $- \boldsymbol{\hat{\mu}}^r_t\sum_{i=1}^N \alpha_i(\textcolor{red}{\boldsymbol{z}^h_t})(\boldsymbol{\hat{\mu}}^r_i)^T + \boldsymbol{\hat{\mu}}^r_t(\boldsymbol{\hat{\mu}}^r_t)^T\sum_{i=1}^N \alpha_i(\textcolor{red}{\boldsymbol{z}^h_t})$ becomes $0$.

\end{document}

%% file: mse_table.tex
\begin{table*}[h!]
    \centering
\vspace{2em}
% \resizebox{0.99\textwidth}{!}{
\begin{tabular}{|c|c|c|c|l|}
\hline

Dataset (units) & Action & MILD~\cite{prasad2023learning} & B\"utepage et al.~\cite{butepage2020imitating} & \hfill MoVEInt \hfill\\ \hline
\multirow{4}{*}{\begin{tabular}{c}HHI\\(B\"utepage et al.~\cite{butepage2020imitating})\\(cm)\end{tabular}} & Hand Wave & 0.788 $\pm$ 1.226 &  4.121 $\pm$ 2.252 & \textbf{0.448 $\pm$ 0.630} \\ \cline{2-5} 
& Handshake & 1.654 $\pm$ 1.549 &  1.181 $\pm$ 0.859 & \textbf{0.196 $\pm$ 0.153}\\ \cline{2-5} 
& Rocket Fistbump & 0.370 $\pm$ 0.682 &  0.544 $\pm$ 1.249 & \textbf{0.123 $\pm$ 0.175}\\ \cline{2-5} 
& Parachute Fistbump & 0.537 $\pm$ 0.579 &  0.977 $\pm$ 1.141 & \textbf{0.314 $\pm$ 0.348} \\ \hline
\multirow{4}{*}{\begin{tabular}{c}HRI-Pepper\\(B\"utepage et al.~\cite{butepage2020imitating})\\(rad)\end{tabular}} & Hand Wave & 0.103 $\pm$ 0.103 & 0.664 $\pm$ 0.277 & \textbf{0.087 $\pm$ 0.089}\\ \cline{2-5} 
 & Handshake & 0.056 $\pm$ 0.041 & 0.184 $\pm$ 0.141 & \textbf{0.015 $\pm$ 0.014}\\ \cline{2-5} 
 & Rocket Fistbump & 0.018 $\pm$ 0.035 &  0.033 $\pm$ 0.045 & \textbf{0.007 $\pm$ 0.015}\\ \cline{2-5} 
 & Parachute Fistbump & 0.088 $\pm$ 0.148 & 0.189 $\pm$ 0.196 & \textbf{0.048 $\pm$ 0.112} \\ \hline
\multirow{4}{*}{\begin{tabular}{c}HRI-Yumi\\(B\"utepage et al.~\cite{butepage2020imitating})\\(rad)\end{tabular}} & Hand Wave & 1.033 $\pm$ 1.204 & 0.225 $\pm$ 0.302 & \textbf{0.147 $\pm$ 0.072} \\ \cline{2-5} 
 & Handshake & 0.068 $\pm$ 0.052 & 0.133 $\pm$ 0.214 & \textbf{0.057 $\pm$ 0.044}\\ \cline{2-5} 
 & Rocket Fistbump & 0.128 $\pm$ 0.071 & 0.147 $\pm$ 0.119 & \textbf{0.093 $\pm$ 0.045} \\ \cline{2-5} 
 & Parachute Fistbump & \textbf{0.028 $\pm$ 0.034} & 0.181 $\pm$ 0.155 & 0.081 $\pm$ 0.082\\ \hline
\multirow{4}{*}{\begin{tabular}{c}HHI\\(NuiSI~\cite{prasad2023learning})\\(cm)\end{tabular}} & Hand Wave & 0.408 $\pm$ 0.538 & 3.168 $\pm$ 3.392 & \textbf{0.298 $\pm$ 0.274} \\ \cline{2-5} 
& Handshake & 0.311 $\pm$ 0.259 & 1.489 $\pm$ 3.327 & \textbf{0.149 $\pm$ 0.120} \\ \cline{2-5} 
& Rocket Fistbump & 1.142 $\pm$ 1.375 & 3.576 $\pm$ 3.082 & \textbf{0.673 $\pm$ 0.679} \\ \cline{2-5} 
& Parachute Fistbump & 0.453 $\pm$ 0.578 & 2.008 $\pm$ 2.024 & \textbf{0.291 $\pm$ 0.199} \\ \hline
\multirow{4}{*}{\begin{tabular}{c}HRI-Pepper\\(NuiSI~\cite{prasad2023learning})\\(rad)\end{tabular}} & Hand Wave & 0.046 $\pm$ 0.059 & 0.057 $\pm$ 0.093 & \textbf{0.044 $\pm$ 0.048} \\ \cline{2-5} 
& Handshake & 0.020 $\pm$ 0.014 & 0.083 $\pm$ 0.075 & \textbf{0.011 $\pm$ 0.008} \\ \cline{2-5} 
& Rocket Fistbump  & 0.077 $\pm$ 0.067 & 0.101 $\pm$ 0.086 & \textbf{0.045 $\pm$ 0.045} \\ \cline{2-5} 
& Parachute Fistbump & 0.022 $\pm$ 0.027 & 0.049 $\pm$ 0.040 & \textbf{0.017 $\pm$ 0.014} \\ \hline
HHI-Handovers & Unimanual & 0.441 $\pm$ 0.280 & 1.133 $\pm$ 0.721 & \textbf{0.441 $\pm$ 0.221}\\ \cline{2-5} 
(Kshirsagar et al.~\cite{kshirsagar2023dataset}) (cm) & Bimanual & 0.869 $\pm$ 0.964 & 0.990 $\pm$ 0.764 & \textbf{0.685 $\pm$ 0.643} \\ \hline

\end{tabular}
% }
\vspace{0.5em}
    \caption{Prediction MSE for robot trajectories after observing the human partner averaged over all joints and timesteps. Results for the HHI scenarios are in cm and for the HRI scenarios are in radians. (Lower is better)}
    \label{tab:pred_mse}
\vspace{-1em}
\end{table*}

%% file: hri-kobo.tex
    \begin{figure*}[h!]
    \centering
    % \begin{subfigure}[b]{0.9\textwidth}
    %      \centering
             \includegraphics[width=0.19\textwidth]{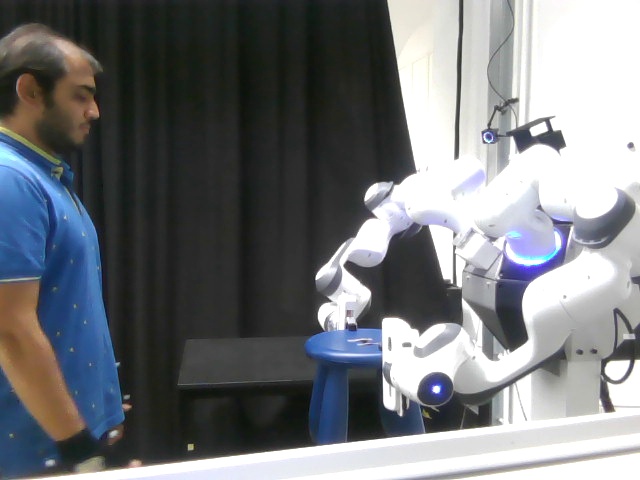}
        \includegraphics[width=0.19\textwidth]{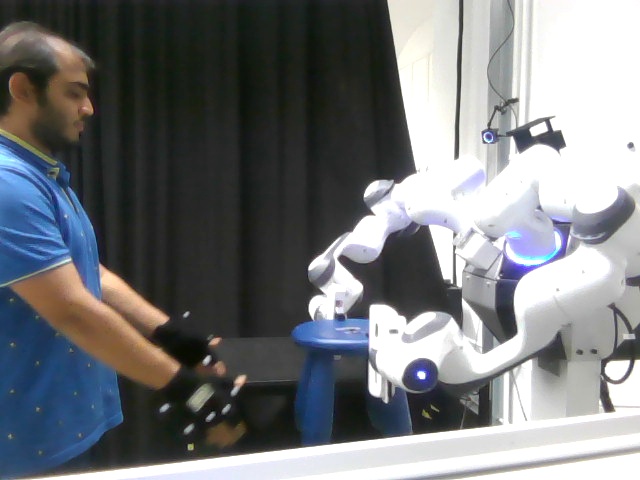}
        \includegraphics[width=0.19\textwidth]{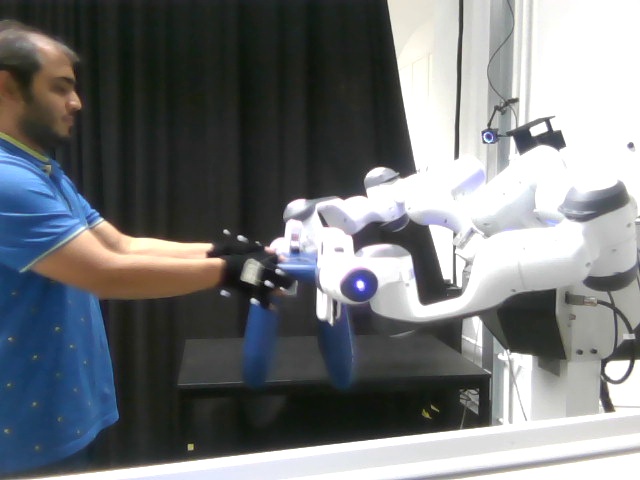}
        \includegraphics[width=0.19\textwidth]{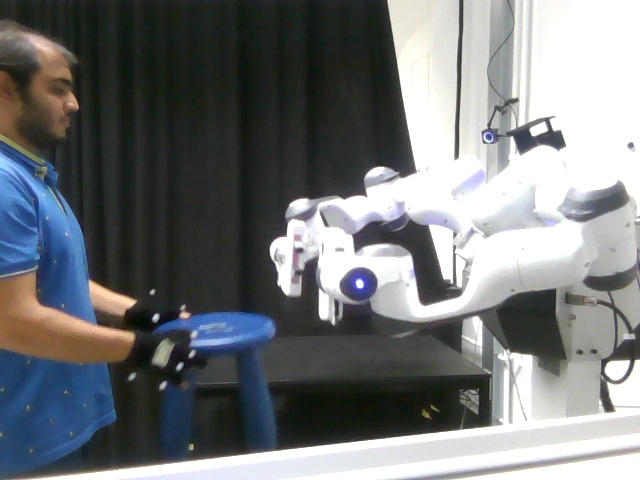}
        \includegraphics[width=0.19\textwidth]{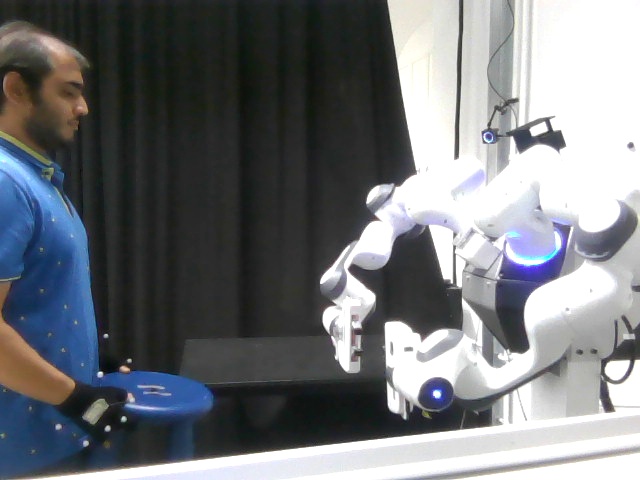}

    \caption{Sample Human-Robot Interactions generated with the reactive motions generated by MoVEInt for a Bimanual Handover scenario.% (Red - Human observation, Dark Blue - Generated Robot Trajectory, Yellow, Purple, and Light Blue - MDN Components Decoded).
    }
    \label{fig:moveint-hri}
\end{figure*}